\documentclass[conference]{IEEEtran}
\IEEEoverridecommandlockouts
\usepackage{cite}
\usepackage{amsmath,amssymb,amsfonts}
\usepackage{graphicx}
\usepackage{textcomp}
\usepackage{xcolor}
\usepackage{pifont}
\usepackage{bbding}
\usepackage{authblk}
\usepackage{xspace}
\usepackage{booktabs}
\usepackage{bbding}
\usepackage{multirow}
\usepackage{algorithm,algpseudocode}
\usepackage{adjustbox}
\usepackage{graphicx}
\usepackage{tikz}
\usetikzlibrary{fit}% 坐标适配-****
\usetikzlibrary{spy}% 局部缩放库
\usetikzlibrary{arrows.meta}% 箭头样式
\usepackage{hyperref}
\definecolor{cvprblue}{rgb}{0.21,0.49,0.74}
\hypersetup{
    colorlinks=true,
    linkcolor=cvprblue, % 链接颜色
    citecolor=cvprblue, % 引用的颜色
    urlcolor=purple % URL颜色
}
\usepackage[capitalize]{cleveref}
\crefname{section}{Sec.}{Secs.}
\Crefname{section}{Section}{Sections}
\Crefname{table}{Table}{Tables}
\crefname{table}{Table}{Tables}
\Crefname{figure}{Figure}{Figures}
\crefname{figure}{Fig.}{Figs.}
\Crefname{equation}{Equation}{Equations}
\crefname{equation}{Eq.}{Eqs.}
\crefname{algocf}{alg.}{algs.}
\Crefname{algocf}{Algorithm}{Algorithms}

\def\BibTeX{{\rm B\kern-.05em{\sc i\kern-.025em b}\kern-.08em
    T\kern-.1667em\lower.7ex\hbox{E}\kern-.125emX}}
\begin{document}

\title{AFFSegNet: Adaptive Feature Fusion Segmentation Network for Microtumors and Multi-Organ Segmentation}

\author[12]{Fuchen Zheng}
\author[4]{Xinyi Chen}
\author[12]{Xuhang Chen}
\author[3]{Haolun Li}
\author[1]{Xiaojiao Guo}
\author[1]{Weihuang Liu}
\author[1*]{\authorcr Chi-Man Pun}
\author[2*]{Shoujun Zhou\thanks{* Corresponding authors.}}
\affil[1]{University of Macau}
\affil[2]{Shenzhen Institute of Advanced Technology, Chinese Academy of Sciences}
\affil[3]{Nanjing University of Posts and Telecommunications}
\affil[4]{Southern University of Science and Technology}

\maketitle

\begin{abstract}
Medical image segmentation, a crucial task in computer vision, facilitates the automated delineation of anatomical structures and pathologies, supporting clinicians in diagnosis, treatment planning, and disease monitoring. However, existing methods are limited in capturing local and global features. To address this limitation, we propose the Adaptive Feature Fusion Segmentation Network (AFFSegNet), a transformer architecture that effectively integrates local and globally features for precise segmentation. Specifically, we introduce an augmented multi-layer perceptron within the encoder to explicitly model long-range dependencies during feature extraction. Furthermore, recognizing the limitations of conventional symmetrical encoder-decoder designs, we present an Enhanced Forward Feedback Network (EFFN) to complement our encoder. Extensive experiments on diverse medical image segmentations, including multi-organ, liver tumor, and bladder tumor, demonstrate the robustness and adaptability of the proposed network across different tumor types and imaging modalities. Finally, we conduct ablation studies to investigate the impact of individual components in the network. These promising results highlight the potential of our proposed network as a robust and valuable tool for assisting medical professionals in critical tasks. Code and models are available at: \url{https://github.com/lzeeorno/AFFSegNet}.
\end{abstract}


\section{Introduction}
% By accurately delineating anatomical structures, segmentation improves the clarity of medical images, significantly improving diagnostic accuracy and efficiency for healthcare professionals. 
Current research in medical image segmentation focuses on critical tasks such as tumor segmentation and organ delineation.
Consequently, neural network architectures from the broader field of computer vision are being increasingly adapted for medical image analysis. Vision transformers~\cite{34vit}, exemplified by the Swin-transformer~\cite{37swin}, have gained significant traction due to their robust feature extraction capabilities. However, while advancements in window attention mechanisms within these transformers have yielded impressive results, challenges remain. These models often struggle to capture the features of small objects due to limitations in modeling long-range dependencies~\cite{79long_range_dependencies2020cluster} and accurately delineating the edges of the image. Consequently, effectively integrating multi-scale local and global features remains an ongoing challenge.

To address these limitations, we propose the Adaptive Feature Fusion Segmentation Network (AFFSegNet), a novel Transformer-based~\cite{15chen2021transunet} architecture specifically designed for medical image segmentation. Inspired by the strengths of ResUnet~\cite{7resnet} and Swin-transformer~\cite{37swin}, AFFSegNet leverages Transformer blocks within a U-shaped residual structure to enhance feature learning across multiple scales.

Furthermore, recognizing the limitations of simply replicating encoder structures in the decoder, we introduce a novel Adaptive Feature Fusion (AFF) decoder. This decoder comprises three key components: the Long Range Dependencies (LRD) block, the Multi-Scale Feature Fusion (MFF) block, and the Adaptive Semantic Center (ASC) block. These components work synergistically to leverage encoder-derived features effectively, enabling the accurate segmentation of small structures, particularly at edges, and facilitating robust multi-scale feature fusion.

Our straightforward network architecture, without relying on complex multi-scale structures or intricate loss functions, achieves state-of-the-art performance on various medical image segmentation tasks. Notably, AFFSegNet surpasses previous state-of-the-art models, demonstrating impressive improvements on the LiTS2017, ISICDM2019 and Synapse datasets, respectively.
The main contributions of this paper are as follows.

1. We introduce \textbf{AFFSegNet}, a hybrid model that combines the strengths of ResUnet and Swin-transformer, incorporating window attention, spatial attention, U-shaped architecture, and residual connections for efficient segmentation.

2. We propose an \textbf{Adaptive Feature Fusion (AFF) Decoder} that maximizes the synergistic potential of window attention to capture multi-scale local and global information by fusing feature maps of varying scales.

3. Extensive experiments demonstrate that the proposed AFFSegNet achieves new state-of-the-art results on various medical image segmentation datasets.

% \begin{figure}[ht]
% \includegraphics[width=1\linewidth,height=0.6\textheight]{figs/PrecSegNet_Overall.pdf}
% \caption{Overview of the PrecSegNet architecture.}
% \label{fig1}
% \end{figure}

\begin{figure*}[ht] 
    \centering
    \includegraphics[width=0.9\linewidth]{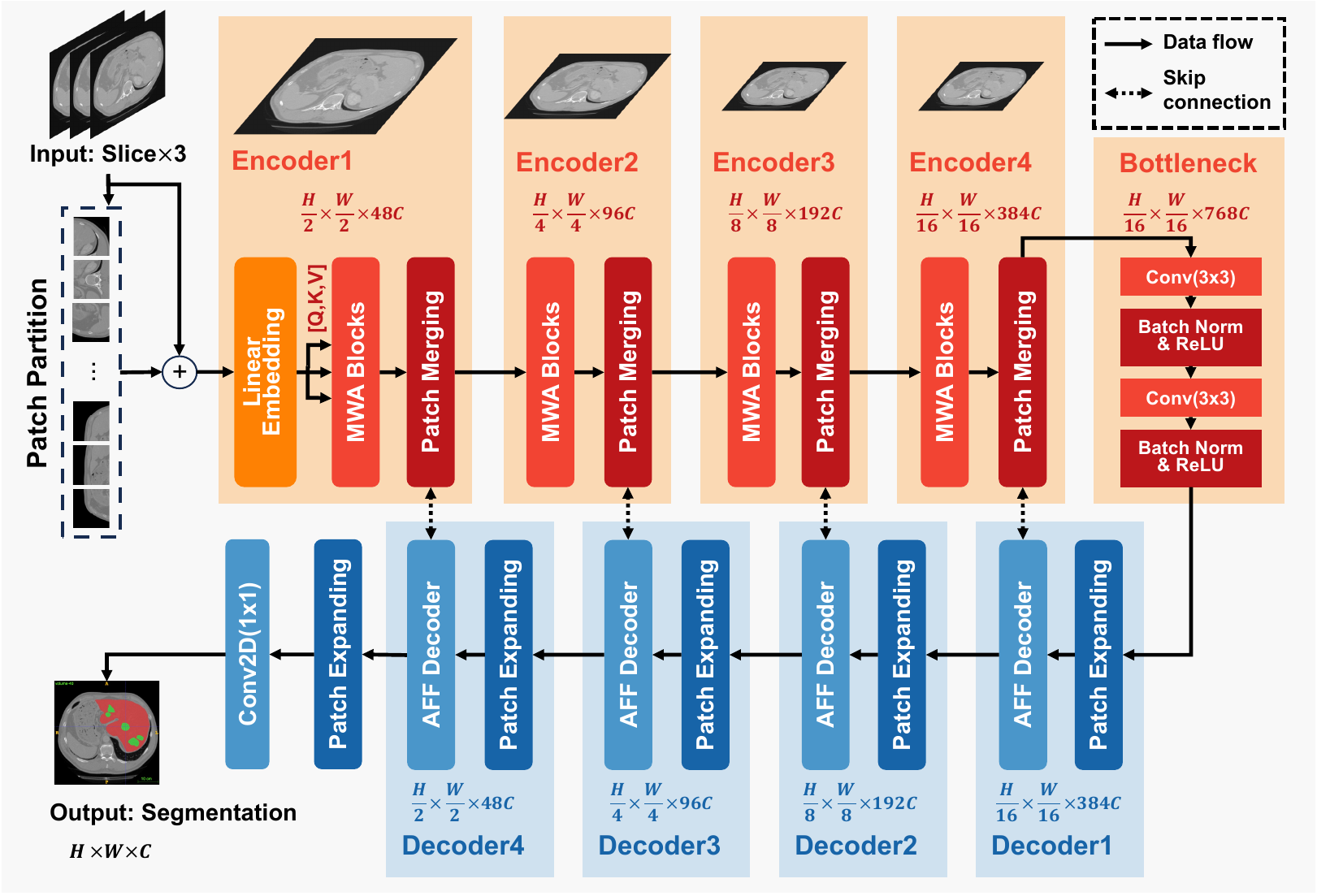}
    \caption{Overview of the AFFSegNet architecture.}
    \label{fig1}
\end{figure*}

\section{Related Work}

% \subsection{Medical Image Segmentation}

% Medical image segmentation is a fundamental task that involves partitioning medical images into distinct regions representing different anatomical structures or tissues. The U-Net architecture~\cite{4unet}, with its elegant encoder-decoder structure and skip connections, has emerged as a leading approach for this task. Its ability to capture both fine-grained details and global context has led to widespread adoption and numerous extensions. For instance, ResUNet~\cite{6resunet} combines the strengths of U-Net~\cite{4unet} and ResNet~\cite{7resnet}, leveraging the power of residual connections~\cite{8residual_connection} to facilitate the training of deeper networks and mitigate the vanishing gradient problem~\cite{9res_learning}. These residual connections allow for unimpeded information flow across layers, significantly enhancing the network's ability to learn complex representations and improve segmentation accuracy. Building upon these foundational concepts, AFFSegNet also incorporates skip connections and residual connections to optimize its segmentation capabilities.

\subsection{Vision Transformer and Hybrid Architectures}

Unlike Convolutional Neural Networks (CNNs) that process images locally, Vision Transformer (ViT) models~\cite{34vit} leverage a self-attention mechanism to capture long-range dependencies within images. This global receptive field has enabled ViT to achieve state-of-the-art performance in image classification tasks. The success of ViT has inspired its adaptation to medical image analysis, with the Swin Transformer~\cite{12swin2021} demonstrating impressive results in various medical imaging applications. The Swin Transformer employs a hierarchical approach, computing self-attention within local windows and then shifting these windows to capture relationships across different image regions. This strategy reduces computational complexity while preserving the ability to model long-range dependencies. However, a common limitation in similar architectures is the suboptimal integration of attention mechanisms, preventing the full realization of the transformer's potential. To address this, our proposed network introduces a novel residual U-shaped transformer architecture designed for effective attention fusion. This architecture leverages the strengths of the window attention mechanism employed in the Swin Transformer and enhances it with an Enhanced Forward Feedback Network (EFFN), resulting in superior performance for medical image segmentation.

\section{Methodology}
This section outlines the architecture and functionality of AFFSegNet. We first describe the network's overall pipeline, followed by a detailed exposition of the Multi-scale Window Attention (MWA) Transformer block, the core encoder component. Subsequently, we elucidate the Adaptive Feature Fusion (AFF) decoder, which is crucial for modeling long-range dependencies and enhancing the network's ability to capture fine-grained details amidst complex edge structures.

\subsection{Overall Pipeline}
AFFSegNet uses a hierarchical U-shaped architecture with skip and residual connections to enhance information flow, as shown in \cref{fig1}.
The input image of size $C\times H\times W$ is processed through patch partitioning and linear embedding before entering the window attention module in the MWA block. Following the Swin Transformer~\cite{12swin2021}, the encoder has four stages, each performing $2\times C$ spatial downsampling in the patch merging layer, which concatenates features from neighboring patches $2\times2$ and applies a linear projection to reduce their dimension.
The decoder mirrors the encoder with four symmetric stages and includes the Adaptive Feature Fusion (AFF) decoder, which combines high-level semantic information with low-level spatial details, outperforming current state-of-the-art models~\cite{12swin2021,21swinUnet2022,49swinUNETR}. An output convolution layer then processes the concatenated features to produce the segmentation prediction.

\subsection{MWA Transformer Block}
\begin{figure}[ht]
\centerline{\includegraphics[width=0.8\columnwidth]{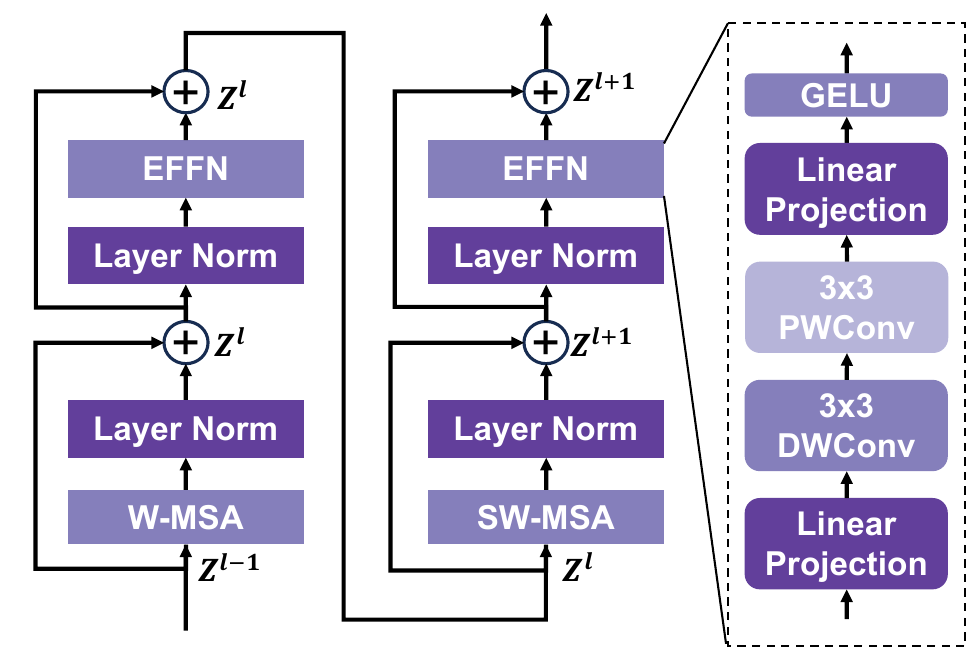}}
\caption{This figure presents details of a schematic diagram of the proposed Multi-scale Window Attention (MWA) transformer block.}
\label{fig2}
\end{figure}

Recognizing the limitations of standard FFNs in capturing local context~\cite{70MLP2021cvt}, we enhance the MLP within our Transformer block by incorporating depth-wise and pixel-wise convolutions~\cite{72depth-wiseConv2021localvit}. As shown in \cref{fig2}, the MWA Transformer block is the backbone of AFFSegNet, which replaces the Multi-Head Self-Attention (MSA) module~\cite{11attention2017} in the standard Transformer with a shifted window attention-based MSA, while keeping other components intact. Each MWA block includes a shifted window-based MSA module followed by an Enhanced Feed-Forward Network (EFFN).

% As depicted in \cref{fig2}, the EFFN first projects input tokens to a higher dimensional space. The projected tokens are then reshaped into 2D feature maps and processed by a $3\times3$ pixel-wise convolution followed by a $3\times3$ depth-wise convolution, effectively capturing local contextual information. Subsequently, the features are reshaped back into tokens and projected back to the original channel dimension. Finally, a GELU activation function~\cite{75GELU2016gaussian} introduces non-linearity.

Mathematically, the computation within an MWA transformer block can be expressed as:
\begin{equation}
\begin{aligned}
& \hat{X}^{l} = \operatorname{W-MSA}(\operatorname{LN}(X^{l-1})) + X^{l-1}, \\
& X^{l} = \operatorname{EFFN}(\operatorname{LN}(\hat{X}^{l})) + \hat{X}^{l},  \\
& \hat{X}^{l+1} = \operatorname{SW-MSA}(\operatorname{LN}(X^{l})) + X^{l}, \\
& X^{l+1} = \operatorname{EFFN}(\operatorname{LN}(\hat{X}^{l+1})) + \hat{X}^{l+1}, 
\end{aligned}
\end{equation}
where $\hat{X}^{l}$ and $\hat{X}^{l+1}$ represent the output from window-based multi-head self-attention using regular ($\operatorname{W-MSA}$) and shifted window partitioning configurations ($\operatorname{SW-MSA}$), respectively; $\operatorname{LN}$ and $\operatorname{EFFN}$ denote layer normalization and the proposed enhanced feed-forward network illustrated in \cref{fig2}, respectively.

Following previous work~\cite{49swinUNETR}, we incorporate a relative position bias $B$ within the self-attention computation to enhance performance. The attention calculation is formulated as:
\begin{equation}
\operatorname{Attention}(Q, K, V) = \operatorname{SoftMax}\left(\frac{QK^T}{\sqrt{d}} + B\right)V,
\end{equation}
where $B$ is derived from a smaller parameterized bias matrix $\hat{B} \in \mathbb{R}^{(2M-1) \times (2M-1)}$; $Q$, $K$, and $V$ represent the query, key, and value matrices, respectively; and $d$ is the dimension of the query and key features.

The synergistic interplay between W-MSA, SW-MSA, and EFFN within each Transformer block enables AFFSegNet to effectively capture both global and local contextual information, leading to improved segmentation performance.

\subsection{Adaptive Feature Fusion (AFF) Decoder}
\begin{figure}[ht]
\centerline{\includegraphics[width=0.8\columnwidth]{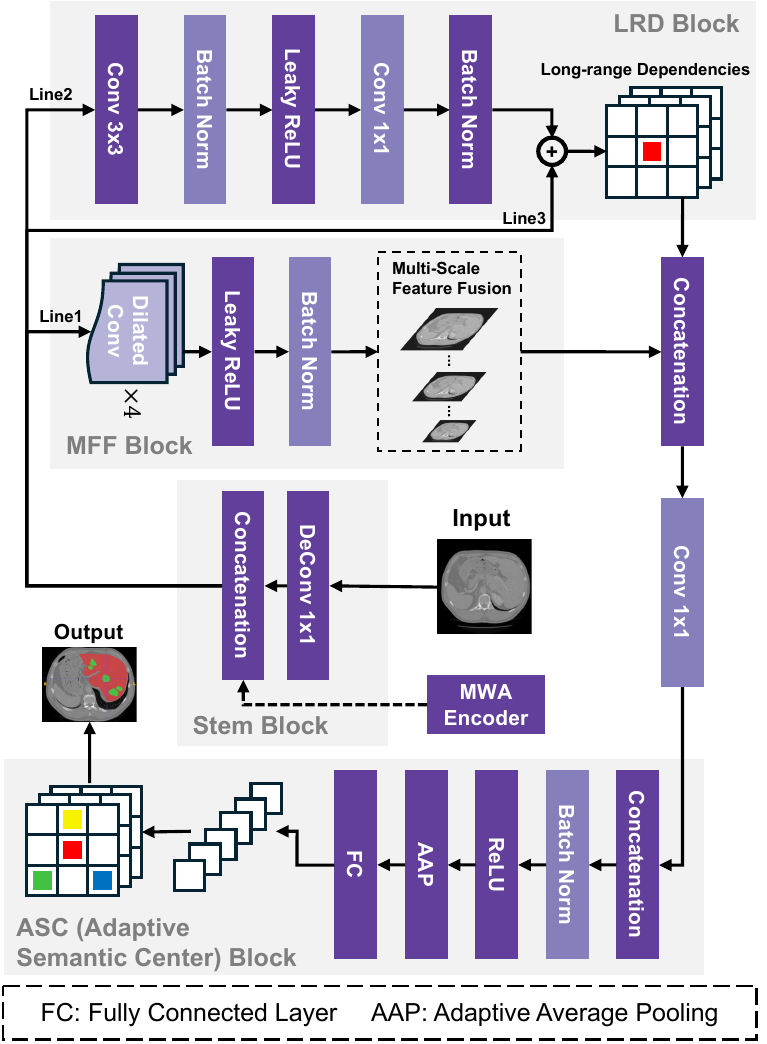}}
\caption{This figure presents details of a schematic diagram of the proposed Adaptive Feature Fusion (AFF) Decoder.}
\label{fig3}
\end{figure}

To address the limitations of vision transformers in capturing local dependencies~\cite{79long_range_dependencies2020cluster} and the inadequacies of existing decoders in integrating multi-scale local and global features~\cite{12swin2021,21swinUnet2022,49swinUNETR}, we propose an Adaptive Feature Fusion (AFF) Decoder. The AFF decoder comprises Long-Range Dependencies (LRD) block, Multi-scale Feature Fusion (MFF) block, and Adaptive Semantic Center (ASC) block, as illustrated in \cref{fig3}.

The AFF decoder begins with a standard deconvolution operation to restore the feature map to the original image size while preserving resolution. Subsequently, skip connections are employed to concatenate MWA encoder feature maps from different scales, enriching the feature map with multi-scale information. This enriched feature map then undergoes three parallel operations. LeakyReLU~\cite{83LeakyReLU2013rectifier} is utilized as the activation function in the decoder to mitigate the vanishing gradient problem and enhance model stability and generalization. LRD block, implemented using a series of convolutions and LeakyReLU activations, models long-range dependencies. Finally, line3 acts as a mask prompt, aiding the decoding process of the line1 and line2 threads. The resulting feature map is then passed to the ASC block. ASC block extracts local region information and performs channel-wise enhancement by utilizing an enhanced filter generated from adaptive average pooling~\cite{85pooling2014mixed} and a fully connected layer~\cite{alexnet2012imagenet}.

\begin{table*}[!ht]
\centering
\caption{Comparison with State-of-the-Art models on the ISICDM2019 and LITS2017 datasets. The best results are bolded while the second best are underlined.}
\begin{tabular}{c|cccc|cccc}
\toprule
\multirow{3}{*}{Method}  & \multicolumn{4}{c|}{ISIDM2019}                                    & \multicolumn{4}{c}{LITS2017}                                      \\ \cmidrule(l){2-9} 
                         & \multicolumn{2}{c}{Average}     & Bladder        & Tumor          & \multicolumn{2}{c}{Average}     & Liver        & Tumor          \\ \cmidrule(l){2-9} 
                         & DSC(\%) $\uparrow$        & mIoU(\%) $\uparrow$       & DSC(\%) $\uparrow$        & DSC(\%) $\uparrow$        & DSC(\%) $\uparrow$        & mIoU(\%) $\uparrow$       & DSC(\%) $\uparrow$        & DSC(\%) $\uparrow$        \\ 
                         \midrule

R50-ViT~\cite{34vit}+CUP~\cite{15chen2021transunet}              & 88.77          & 85.62          & 92.05          & 85.49          & 82.62          & 79.68          & 85.83          & 79.41          \\

TransUNet~\cite{15chen2021transunet}                
& \underline{94.56} & \underline{93.60}  & \underline{97.74} & \underline{91.38} & \underline{93.29} & \underline{90.81}  & \underline{95.54} & \underline{91.03}               \\

SwinUNet~\cite{21swinUnet2022}                 
& 91.95 & 89.77 & 94.73 & 89.17 & 89.68 & 86.62 & 93.31 & 86.04               \\

Swin UNETR~\cite{49swinUNETR}
& 92.60 & 90.61 & 95.08 & 90.12 & 91.95 & 90.02 & 94.73 & 89.17               \\

UNETR~\cite{48unetr}                    & 91.55          & 88.34          & 94.83          & 88.26          &  89.38 & 87.46 & 92.89 & 85.86               \\

nnFormer~\cite{47nnformer}                 & 93.69          & 89.11          & 96.97          & 90.41          &    91.74 & 89.95 & 94.57 & 88.91              \\ 

\midrule

SAM~\cite{87sam_meta2023segment}+Point Prompt                & 34.16          & 23.4          & 59.10          & 9.22          &    27.33 & 17.21 & 46.10 & 8.56              \\

% MedSAM~\cite{89medsam_nature2024segment}                 & 53.17          & 37.16          & 65.3          & 41.03          & 73.36    & 60.73 & 90.1 & 56.61              \\ 

\midrule
\textbf{AFFSegNet (Ours)} & \textbf{96.75} & \textbf{96.04} & \textbf{98.87} & \textbf{94.63} & \textbf{95.47} & \textbf{94.88} & \textbf{96.79} & \textbf{94.14} \\ 

% \midrule

% P-values & $< 1e^{-2} \text{(DSC)}$, $< 1e^{-2} \text{(mIoU)}$
% P-values & \multicolumn{8}{c}{$< 1e^{-2} \text{(DSC)}$, $< 1e^{-2}\text{(mIoU)}$} \\

\bottomrule
\end{tabular}
\label{tab:comp}
\end{table*}

\begin{table*}[ht]
    \centering
    \caption{Comparison with State-of-the-Art models on the Synapse multi-organ dataset. The best results are bolded while the second best are underlined.
    }
    \adjustbox{width=\linewidth}{
    % \resizebox{\textwidth}{!}{%
    \begin{tabular}{c|c|c|c|c|c|c|c|c|c}
        \toprule
        Model  & \multicolumn{1}{c|}{Average} & Aotra & Gallbladder & Kidney(Left) & Kidney(Right) & Liver & Pancreas & Spleen & Stomach \\
               & DSC(\%)$\uparrow$ & DSC(\%)$\uparrow$ & DSC(\%)$\uparrow$ & DSC(\%)$\uparrow$ & DSC(\%)$\uparrow$ & DSC(\%)$\uparrow$ & DSC(\%)$\uparrow$ & DSC(\%)$\uparrow$ & DSC(\%)$\uparrow$\\
        \midrule

        %following result from nnformer
        % ViT~\cite{34vit}+CUP~\cite{15chen2021transunet}    
        %     & 67.86 & 70.19 & 45.10 & 74.70 & 67.40 & 91.32 & 42.00 & 81.75 & 70.44     \\
    
        R50-ViT~\cite{34vit}+CUP~\cite{15chen2021transunet}  
            & 71.29 & 73.73 & 55.13 & 75.80 & 72.20 & 91.51 & 45.99 & 81.99 & 73.95     \\
            
        TransUNet~\cite{15chen2021transunet}
            & 84.37 & 90.68 & 71.99 & 86.04 & 83.71 & 95.54 & 73.96 & 88.80 & 84.20     \\
            
        SwinUNet~\cite{21swinUnet2022}
            & 79.13 & 85.47 & 66.53 & 83.28 & 79.61 & 94.29 & 56.58 & 90.66 & 76.60    \\

        UNETR~\cite{48unetr}
            & 79.57 & 89.99 & 60.56 & 85.66 & 84.80 & 94.46 & 59.25 & 87.81 & 73.99     \\

        % TransClaw U-Net~\cite{18chang2021transclaw}
        %     & 78.09 & 85.87 & 61.38 & 84.83 & 79.36 & 94.28 & 57.65 & 87.74 & 73.55    \\

        % LeViT-UNet-384s~\cite{50levit}
        %     & 78.53 & 87.33 & 62.23 & 84.61 & 80.25 & 93.11 & 59.07 & 88.86 & 72.76     \\

        % MISSFormer~\cite{51missformer}
        %     & 81.96 & 86.99 & 68.65 & 85.21 & 82.00 & 94.41 & 65.67 & 91.92 & 80.81    \\
        
        %following result from 3D Transnet   
        Swin UNETR~\cite{49swinUNETR}
                    & 83.51 & \underline{90.75} & 66.72 & 86.51 & 85.88 & 95.33 & 70.07 & \textbf{94.59} & 78.20     \\

        nnFormer~\cite{47nnformer}
            & 85.32 & 90.72 & 71.67 & 85.60 & 87.02 &  \underline{96.28} & 82.28 &  87.30 & 81.69     \\

        \midrule

        SAM~\cite{87sam_meta2023segment}+Point Prompt
        & 58.55 & 61.20 & 54.30 & 79.10 & 68.60 &  46.10 & 51.10 &  51.80 & 56.20     \\

        MedSAM~\cite{89medsam_nature2024segment}
        & 82.55 & 87.20 & 76.60 & 88.50 & 81.40 &  90.10 & 76.00 &  75.10 & 85.50     \\

        SAM 2~\cite{92SAM2meta2024sam}
        & 53.39 & 40.00 & 77.20 & 64.20 & 72.40 & 27.00  & 68.20 & 36.60  & 41.50     \\

        MedSAM-2~\cite{90medsam2_2024medical}
        & \underline{89.08} & 89.40 & \textbf{92.70} & \underline{92.10} & \underline{92.40} & 83.60  & \textbf{83.20} & 91.80  & \underline{87.40}     \\
        
        \midrule
        %t= R5-0.5, B= unetr - 0.5
        \textbf{AFFSegNet (Ours)}
            & \textbf{90.73} & \textbf{93.02} & \underline{87.08} & \textbf{92.67} & \textbf{93.06} & \textbf{97.11} & \underline{82.97} & \underline{92.19} & \textbf{87.72}     \\

        \bottomrule
    \end{tabular}
    }
    \label{tab:syna}
\end{table*}

\subsection{Objective Function}
During training, AFFSegNet employs the BCE Dice loss $\mathcal{L}_{BD}$~\cite{62BCEDiceLoss2016v}, a combination of Binary Cross-Entropy (BCE) loss $\mathcal{L}_{BCE}$ and Dice loss $\mathcal{L}_{D}$, widely used in medical image segmentation tasks. This loss function is defined as:
\begin{equation}
\begin{aligned}
    \mathcal{L}_{BD} = & \mathcal{L}_{D} + \mathcal{L}_{BCE}(y, p) \\
    = & \frac{1}{N} \sum_{i=1}^{N} \left(1 - \frac{2 \sum_{j} y_{i,j} p_{i,j}}{\sum_{j} y_{i,j} + \sum_{j} p_{i,j}}\right) \\ & -{(y\log(p) + (1 - y)\log(1 - p))},
\end{aligned}
\end{equation}
where $y$ represents the ground truth segmentation mask, $p$ denotes the predicted segmentation mask, and $N$ is the number of pixels in the image.

\section{Experiments}
In this section, we present the experimental framework and discuss the results. First, we describe the datasets and evaluation metrics employed. Next, we compare the performance of AFFSegNet against state-of-the-art methods in medical image segmentation. Finally, we conduct ablation studies to investigate the impact of individual components in the AFFSegNet architecture.

\subsection{Datasets and Implementation Details}\label{sec:4a}
To ensure a comprehensive evaluation and fair comparison with existing methods, experiments were conducted on three public medical image datasets. 
1. LiTS2017~\cite{44lits2017}: This dataset focuses on liver tumor segmentation and comprises 131 contrast-enhanced 3D abdominal CT scans.
2. ISICDM2019~\cite{46ISICDM2019}: This dataset centers on bladder tumor segmentation and includes 2200 bladder cancer images.
3. Synapse~\cite{45synapse}: This dataset targets multi-organ segmentation and consists of 40 3D abdominal CT scans with multiple organs.
    
% \begin{enumerate}
%     \item \textbf{LiTS2017}~\cite{44lits2017}: This dataset focuses on liver tumor segmentation and comprises 131 contrast-enhanced 3D abdominal CT scans.
%     \item \textbf{ISICDM2019}~\cite{46ISICDM2019}: This dataset centers on bladder tumor segmentation and includes 2200 bladder cancer images.
%     \item \textbf{Synapse}~\cite{45synapse}: This dataset targets multi-organ segmentation and consists of 40 3D abdominal CT scans with multiple organs.
% \end{enumerate}

In all experiments, we utilized the nnformer~\cite{47nnformer} dataset splits (80\% training, 15\% validation, 5\% testing) for consistency and fair comparison. Input images were resized to $512 \times 512$ pixels. AFFSegNet was implemented in PyTorch and trained on an NVIDIA GeForce RTX 4090 GPU. We used the SGD optimizer~\cite{63SGD2011adaptive} with a momentum of 0.98, weight decay of $1 \times 10^{-6}$, and an initial learning rate of $1 \times 10^{-2}$, reduced via cosine decay to $6 \times 10^{-6}$. Data augmentation included random horizontal flipping and rotation. Certain experimental results that contradict established common sense are referenced are referenced from nnformer~\cite{47nnformer}, TransUNet~\cite{15chen2021transunet}, SAM~\cite{88medsam_liver20233dsam,91medLSAM_Bladder2023medlsam}, and MedSAM2~\cite{90medsam2_2024medical}.

% \subsection{Evaluation Metrics} 
% We assessed the segmentation performance using two widely recognized metrics:
% \subsubsection{Dice Coefficient Score~\cite{64dice1945measures}}
% The Dice Similarity Coefficient (DSC) quantifies the overlap between the predicted segmentation and the ground truth. Its value ranges from 0 to 1, with higher values indicating superior segmentation performance.
% \begin{equation}
%     DSC = \frac{2 \times |P \cap G|}{|P| + |G|},
% \end{equation}
% In this context, $ P $ denotes the predicted segmentation, while $ G $ represents the ground truth segmentation. The term $ |P \cap G| $ indicates the number of pixels in the intersection of the predicted and ground truth segmentations, whereas $ |P| $ and $ |G| $ denote the number of pixels in the predicted and ground truth segmentations, respectively.

% \subsubsection{Mean Intersection over Union (mIoU)}
% The mean Intersection over Union (mIoU) calculates the average ratio of intersection to union between the predicted segmentation and the ground truth across all classes. It is defined as:
% \begin{equation}
%     mIoU = \frac{1}{C} \sum_{i=1}^{C} \frac{|P_i \cap G_i|}{|P_i| + |G_i| - |P_i \cap G_i|},
% \end{equation}
% where $C$ is the number of classes, $P_i$ represents the predicted segmentation for class $i$, and $G_i$ represents the ground truth segmentation for class $i$. Similar to DSC, mIoU ranges from 0 to 1, with higher values indicating better segmentation performance.
\begin{figure*}[ht]
\centerline{\includegraphics[width=1\linewidth]{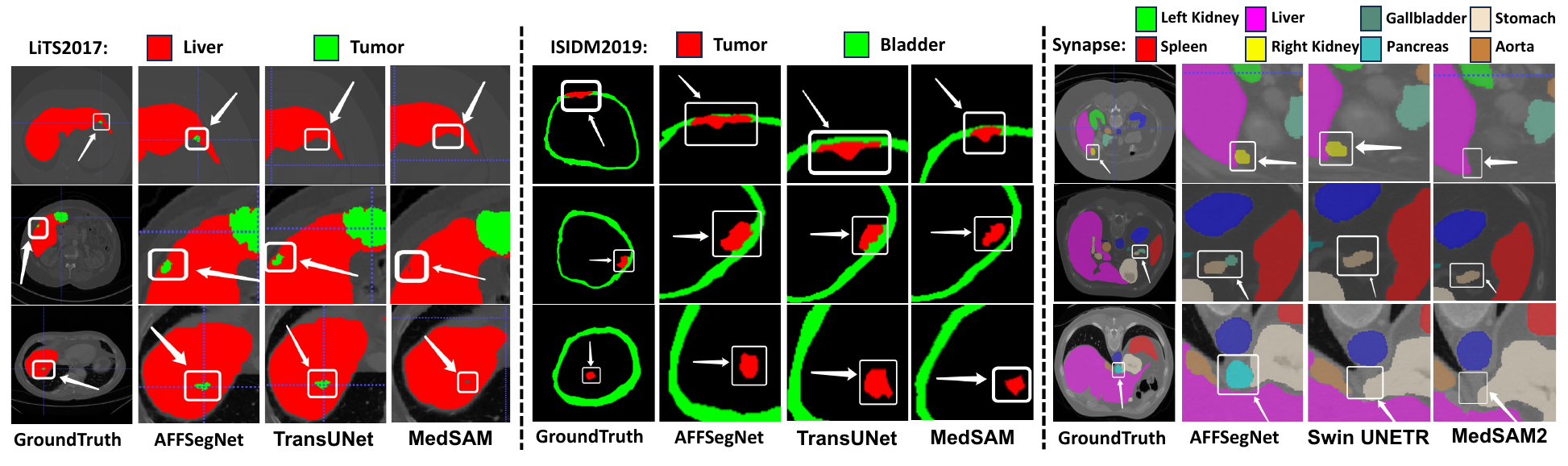}}
\caption{LiTS2017, ISICDM2019 and Synapse Prediction Results}
\label{fig4}
\end{figure*}

We evaluated segmentation performance using two widely recognized metrics:
\subsubsection{Dice Coefficient Score}
The Dice Similarity Coefficient (DSC)~\cite{64dice1945measures} quantifies the overlap between predicted segmentation and ground truth.
\subsubsection{Mean Intersection over Union (mIoU)}
The mean Intersection over Union (mIoU)~\cite{76IoU2010pascal} calculates the average ratio of intersection to union between predicted segmentation and ground truth across all classes.

\subsection{Comparisons with State-of-the-Art Methods}
We compared the performance of AFFSegNet with several state-of-the-art medical image segmentation methods on the three datasets described above. The results are summarized in \cref{tab:comp,tab:syna}.

\subsubsection{Liver Tumor Segmentation}
\Cref{tab:comp} presents the results on the LiTS2017~\cite{44lits2017} dataset. AFFSegNet outperforms all other methods, achieving an average DSC of 95.47\% and an mIoU of 94.88\%. In particular, AFFSegNet surpasses the second-best model, TransUNet~\cite{15chen2021transunet}, by a significant margin (DSC: +2.19\%, mIoU: +4.07\%). This improvement highlights AFFSegNet's ability to accurately segment small and irregularly shaped tumors, which can be attributed to the Multi-Scale Feature Fusion (MFF) block within the AFF Decoder. The MFF block effectively captures features across multiple scales, enabling the network to delineate fine-grained tumor boundaries. Interestingly, the popular segmentation models SAM~\cite{87sam_meta2023segment, 88medsam_liver20233dsam} and MedSAM~\cite{89medsam_nature2024segment} struggle to accurately segment multiple tumors with varying shapes and sizes within a single image even they pretrained on large datasets. This suggests that AFFSegNet's architectural advantages provide it with an edge in handling such complex segmentation scenarios.

\subsubsection{Bladder Tumor Segmentation}
On the ISICDM2019~\cite{46ISICDM2019} dataset, AFFSegNet again demonstrates superior performance, achieving an average DSC of 96.75\% and an mIoU of 96.04\% as shown in \cref{tab:comp}. This represents a substantial improvement of 3.25\% in DSC compared to the second-best method. The remarkable performance on bladder tumor segmentation can be attributed to the ASC block in the AFF decoder, which effectively captures local region information critical for accurate boundary delineation. These results underscore the robustness and adaptability of AFFSegNet across different tumor types and imaging modalities.

\subsubsection{Multi-Organ Segmentation}
The results for the Synapse~\cite{45synapse} multi-organ segmentation dataset are presented in \cref{tab:syna}. AFFSegNet achieves state-of-the-art results with an average DSC of 90 73\%, and AFFSegNet consistently achieves high scores in all organs, demonstrating its ability to generalize to different anatomical structures. In particular, AFFSegNet excels in segmenting smaller organs, achieving the highest DSC scores for five out of the eight organs. This robust performance on a challenging multi-organ dataset highlights the effectiveness of AFFSegNet's U-shaped architecture and AFF decoder in preserving both high-level semantic information and low-level spatial details.

\subsection{Ablation Study}
\begin{center}
\begin{table}[h]
    \centering
    \caption{Ablation study of different modules in ASSNet.}
    % \resizebox{\columnwidth}{!}{
    \begin{tabular}{c c c c|c|c}
        \toprule
        EFFN  & LRD  & MFF & ASC &
        ISICDM2019 & 
        LiTS2017 \\
         &  &  &  & Average DSC $\uparrow$ & Average DSC $\uparrow$ \\
        
        \midrule
    
            $\times$ & \checkmark & \checkmark & \checkmark & 93.91\% & 92.56\%\\
        
            \checkmark & $\times$ & \checkmark & \checkmark & 75.54\% & 73.92\% \\
        
            \checkmark & \checkmark & $\times$ & \checkmark & 87.15\% & 85.10\% \\

            \checkmark & \checkmark & \checkmark & $\times$ & 88.93\% & 87.22\% \\

            % $\times$ & \checkmark & $\times$ & \checkmark & 86.31\% & 84.26\% \\

            \checkmark & \checkmark & \checkmark & \checkmark & 96.75\% & 95.47\% \\
        \bottomrule
    \end{tabular}
    % }
    \label{tab:ablation}
\end{table}
\end{center}

To investigate the contribution of each module within AFFSegNet, we conducted an ablation study on the ISICDM2019 and LiTS2017 datasets. We used the same experimental setup as described in \cref{sec:4a} and evaluated the performance of AFFSegNet by removing one component at a time. The results, summarized in \cref{tab:ablation}, demonstrate that all components contribute to the overall performance of AFFSegNet.
The ablation study clearly shows that the Embedded Feature Fusion Network (EFFN) significantly enhances AFFSegNet's ability to model long-range dependencies. Removing EFFN leads to a considerable drop in performance. This highlights the importance of EFFN in the capture of long-range interactions between image regions.
Similarly, the LRD block in the AFF decoder plays a crucial role in preserving long-range dependencies and establishing a connection between the encoder and the decoder. Removing the LRD block results in a substantial decline in performance, with average DSCs dropping to 75.54\% and 73.92\% for the two datasets, respectively. This confirms the essential function of the LRD block.
The MFF and ASC blocks within the decoder also contribute significantly to AFFSegNet's state-of-the-art performance. Removing the MFF block leads to a decrease in the average DSC to 87.15\% and 85.10\% for the two datasets, respectively, demonstrating the importance of multi-scale feature fusion in medical image segmentation. The ASC block, on the other hand, focuses on detecting critical edges and central features, which are essential for accurate boundary delineation.

\section{Visualization of Segmentation Results}

To visually assess the segmentation capabilities of AFFSegNet, \cref{fig4} presents qualitative comparisons against other state-of-the-art methods in representative slices from the LiTS2017, ISICDM2019 and Synapse datasets. AFFSegNet accurately segments small tumor nodules in the periphery of the liver and preserves the integrity of miniature organs, which are often missed or inaccurately segmented by other methods. These visual comparisons further emphasize the accuracy and robustness of AFFSegNet in challenging medical image segmentation scenarios.

\section{Conclusion}
% The hallmark of AFFSegNet lies in its innovatively and effectively integrating an EFFN, a LRD block, a MFF block, and an ASC block, so that it exhibits superior capacity in modeling long-range dependencies and capturing salient features across varying scales and spatial contexts. Rigorous evaluation on three publicly available medical image segmentation datasets demonstrated that AFFSegNet consistently achieves state-of-the-art segmentation performance, surpassing existing methods in accurately delineating diverse anatomical structures and pathologies. These promising results highlight AFFSegNet's potential as a robust and valuable tool for assisting medical professionals in critical tasks such as diagnosis, treatment planning, and disease monitoring. Future research will investigate the generalizability of AFFSegNet to other medical imaging modalities and further evaluate its performance in more complex clinical scenarios.
This paper presents AFFSegNet, a novel Transformer-based architecture tailored for medical image segmentation tasks, particularly excelling in microtumor and multi-organ segmentation. By integrating an augmented multi-layer perceptron in the encoder and introducing the Adaptive Feature Fusion (AFF) decoder, comprising the Long-Range Dependencies block, Multi-Scale Feature Fusion block, and Adaptive Semantic Center block, AFFSegNet effectively captures both local and global features across multiple scales. The extensive experiments conducted on diverse datasets, including LiTS2017, ISICDM2019, and Synapse, demonstrate that AFFSegNet consistently outperforms existing state-of-the-art models, achieving higher Dice Similarity Coefficients and mIoU scores. Ablation studies further validate the significance of each component within the architecture, underscoring their collective contribution to the network's superior performance. These findings highlight the potential of AFFSegNet as a robust and valuable tool to enhance the precision and efficiency of medical image segmentation, thus supporting clinicians in critical diagnostic and treatment planning processes.

\bibliographystyle{IEEEtran}
\bibliography{ref}

\begin{thebibliography}{10}
\providecommand{\url}[1]{#1}
\csname url@samestyle\endcsname
\providecommand{\newblock}{\relax}
\providecommand{\bibinfo}[2]{#2}
\providecommand{\BIBentrySTDinterwordspacing}{\spaceskip=0pt\relax}
\providecommand{\BIBentryALTinterwordstretchfactor}{4}
\providecommand{\BIBentryALTinterwordspacing}{\spaceskip=\fontdimen2\font plus
\BIBentryALTinterwordstretchfactor\fontdimen3\font minus \fontdimen4\font\relax}
\providecommand{\BIBforeignlanguage}[2]{{%
\expandafter\ifx\csname l@#1\endcsname\relax
\typeout{** WARNING: IEEEtran.bst: No hyphenation pattern has been}%
\typeout{** loaded for the language `#1'. Using the pattern for}%
\typeout{** the default language instead.}%
\else
\language=\csname l@#1\endcsname
\fi
#2}}
\providecommand{\BIBdecl}{\relax}
\BIBdecl

\bibitem{34vit}
A.~Dosovitskiy, L.~Beyer, A.~Kolesnikov, D.~Weissenborn, X.~Zhai, T.~Unterthiner, M.~Dehghani, M.~Minderer, G.~Heigold, S.~Gelly, J.~Uszkoreit, and N.~Houlsby, ``An image is worth 16x16 words: Transformers for image recognition at scale,'' in \emph{ICLR}, 2021.

\bibitem{37swin}
Z.~Liu, Y.~Lin, Y.~Cao, H.~Hu, Y.~Wei, Z.~Zhang, S.~Lin, and B.~Guo, ``Swin transformer: Hierarchical vision transformer using shifted windows,'' in \emph{ICCV}, 2021, pp. 9992--10\,002.

\bibitem{79long_range_dependencies2020cluster}
S.~Wang, L.~Zhou, Z.~Gan, Y.-C. Chen, Y.~Fang, S.~Sun, Y.~Cheng, and J.~Liu, ``Cluster-former: Clustering-based sparse transformer for long-range dependency encoding,'' \emph{arXiv}, 2020.

\bibitem{15chen2021transunet}
J.~Chen, Y.~Lu, Q.~Yu, X.~Luo, E.~Adeli, Y.~Wang, L.~Lu, A.~L. Yuille, and Y.~Zhou, ``Transunet: Transformers make strong encoders for medical image segmentation,'' \emph{ArXiv}, 2021.

\bibitem{7resnet}
K.~He, X.~Zhang, S.~Ren, and J.~Sun, ``Deep residual learning for image recognition,'' in \emph{CVPR}, 2016, pp. 770--778.

\bibitem{12swin2021}
Z.~Liu, Y.~Lin, Y.~Cao, H.~Hu, Y.~Wei, Z.~Zhang, S.~Lin, and B.~Guo, ``Swin transformer: Hierarchical vision transformer using shifted windows,'' in \emph{ICCV}, 2021, pp. 9992--10\,002.

\bibitem{21swinUnet2022}
H.~Cao, Y.~Wang, J.~Chen, D.~Jiang, X.~Zhang, Q.~Tian, and M.~Wang, ``Swin-unet: Unet-like pure transformer for medical image segmentation,'' in \emph{ECCV}, 2022, pp. 205--218.

\bibitem{49swinUNETR}
Y.~Tang, D.~Yang, W.~Li, H.~R. Roth, B.~Landman, D.~Xu, V.~Nath, and A.~Hatamizadeh, ``Self-supervised pre-training of swin transformers for 3d mia,'' in \emph{CVPR}, 2022, pp. 20\,730--20\,740.

\bibitem{70MLP2021cvt}
H.~Wu, B.~Xiao, N.~Codella, M.~Liu, X.~Dai, L.~Yuan, and L.~Zhang, ``Cvt: Introducing convolutions to vision transformers,'' in \emph{ICCV}, 2021, pp. 22--31.

\bibitem{72depth-wiseConv2021localvit}
Y.~Li, K.~Zhang, J.~Cao, R.~Timofte, and L.~Van~Gool, ``Localvit: Bringing locality to vision transformers,'' \emph{arXiv}, 2021.

\bibitem{11attention2017}
A.~Vaswani, N.~Shazeer, N.~Parmar, J.~Uszkoreit, L.~Jones, A.~N. Gomez, L.~Kaiser, and I.~Polosukhin, ``Attention is all you need,'' in \emph{NeurIPS}, 2017, pp. 5998--6008.

\bibitem{83LeakyReLU2013rectifier}
A.~L. Maas, A.~Y. Hannun, A.~Y. Ng \emph{et~al.}, ``Rectifier nonlinearities improve neural network acoustic models,'' in \emph{ICML}, vol.~30, 2013.

\bibitem{85pooling2014mixed}
D.~Yu, H.~Wang, P.~Chen, and Z.~Wei, ``Mixed pooling for convolutional neural networks,'' in \emph{RSKT 2014, Proceedings 9}, 2014, pp. 364--375.

\bibitem{alexnet2012imagenet}
A.~Krizhevsky, I.~Sutskever, and G.~E. Hinton, ``Imagenet classification with deep convolutional neural networks,'' \emph{NeurIPS}, vol.~25, 2012.

\bibitem{48unetr}
A.~Hatamizadeh, Y.~Tang, V.~Nath, D.~Yang, A.~Myronenko, B.~Landman, H.~R. Roth, and D.~Xu, ``Unetr: Transformers for 3d medical image segmentation,'' in \emph{ICCV}, 2022, pp. 574--584.

\bibitem{47nnformer}
H.-Y. Zhou, J.~Guo, Y.~Zhang, X.~Han, L.~Yu, L.~Wang, and Y.~Yu, ``nnformer: Volumetric medical image segmentation via a 3d transformer,'' \emph{IEEE Transactions on Image Processing}, 2023.

\bibitem{87sam_meta2023segment}
A.~Kirillov, E.~Mintun, N.~Ravi, H.~Mao, C.~Rolland, L.~Gustafson, T.~Xiao, S.~Whitehead, A.~C. Berg, W.-Y. Lo \emph{et~al.}, ``Segment anything,'' in \emph{ICCV}, 2023, pp. 4015--4026.

\bibitem{89medsam_nature2024segment}
J.~Ma, Y.~He, F.~Li, L.~Han, C.~You, and B.~Wang, ``Segment anything in medical images,'' \emph{Nature Communications}, vol.~15, no.~1, p. 654, 2024.

\bibitem{92SAM2meta2024sam}
N.~Ravi, V.~Gabeur, Y.-T. Hu, R.~Hu, C.~Ryali, T.~Ma, H.~Khedr, R.~R{\"a}dle, C.~Rolland, L.~Gustafson \emph{et~al.}, ``Sam 2: Segment anything in images and videos,'' \emph{arXiv}, 2024.

\bibitem{90medsam2_2024medical}
J.~Zhu, Y.~Qi, and J.~Wu, ``Medical sam 2: Segment medical images as video via segment anything model 2,'' \emph{arXiv}, 2024.

\bibitem{62BCEDiceLoss2016v}
F.~Milletari, N.~Navab, and S.-A. Ahmadi, ``V-net: Fully convolutional neural networks for volumetric medical image segmentation,'' in \emph{3DV}, 2016, pp. 565--571.

\bibitem{44lits2017}
P.~Bilic, P.~Christ, H.~B. Li, E.~Vorontsov, A.~Ben-Cohen, G.~Kaissis, A.~Szeskin, C.~Jacobs, G.~E.~H. Mamani, G.~Chartrand \emph{et~al.}, ``The liver tumor segmentation benchmark (lits),'' \emph{MIA}, vol.~84, p. 102680, 2023.

\bibitem{46ISICDM2019}
\emph{Proceedings of the Third International Symposium on Image Computing and Digital Medicine, {ISICDM} 2019, Xi'an, China}, 2019.

\bibitem{45synapse}
B.~Landman, Z.~Xu, J.~Igelsias, M.~Styner, T.~Langerak, and A.~Klein, ``Miccai multi-atlas labeling beyond the cranial vault--workshop and challenge,'' in \emph{MICCAI}, vol.~5, 2015, p.~12.

\bibitem{63SGD2011adaptive}
J.~Duchi, E.~Hazan, and Y.~Singer, ``Adaptive subgradient methods for online learning and stochastic optimization.'' \emph{JMLR}, vol.~12, no.~7, 2011.

\bibitem{88medsam_liver20233dsam}
S.~Gong, Y.~Zhong, W.~Ma, J.~Li, Z.~Wang, J.~Zhang, P.-A. Heng, and Q.~Dou, ``3dsam-adapter: Holistic adaptation of sam from 2d to 3d for promptable medical image segmentation,'' \emph{arXiv}, 2023.

\bibitem{91medLSAM_Bladder2023medlsam}
W.~Lei, X.~Wei, X.~Zhang, K.~Li, and S.~Zhang, ``Medlsam: Localize and segment anything model for 3d medical images,'' \emph{arXiv}, 2023.

\bibitem{64dice1945measures}
L.~R. Dice, ``Measures of the amount of ecologic association between species,'' \emph{Ecology}, vol.~26, no.~3, pp. 297--302, 1945.

\bibitem{76IoU2010pascal}
M.~Everingham, L.~Van~Gool, C.~K. Williams, J.~Winn, and A.~Zisserman, ``The pascal visual object classes (voc) challenge,'' \emph{IJCV}, vol.~88, pp. 303--338, 2010.

\end{thebibliography}

\end{document}